\documentclass[pdflatex,sn-mathphys-num]{sn-jnl}

\usepackage{graphicx}%
\usepackage{multirow}%
\usepackage{amsmath,amssymb,amsfonts}%
\usepackage{amsthm}%
\usepackage{mathrsfs}%
\usepackage[title]{appendix}%
\usepackage{xcolor}%
\usepackage{textcomp}%
\usepackage{manyfoot}%
\usepackage{booktabs}%
\usepackage{algorithm}%
\usepackage{algorithmicx}%
\usepackage{algpseudocode}%
\usepackage{listings}%
\usepackage{bbm}
\usepackage{graphicx}  
\usepackage{subcaption} 
\usepackage{todonotes}

\theoremstyle{thmstyleone}%
\theoremstyle{thmstyletwo}%

\theoremstyle{thmstylethree}%

\begin{document}

\title[Privacy-Preserving Federated Unsupervised Domain Adaptation for Regression on Small-Scale and High-Dimensional Biological Data]{Privacy-Preserving Federated Unsupervised Domain Adaptation for Regression on Small-Scale and High-Dimensional Biological Data}

\author*[1,2]{\fnm{Cem Ata} \sur{Baykara}}\email{cem.baykara@uni-tuebingen.de}

\author[1,2]{\fnm{Ali Burak} \sur{Ünal}}

\author[2]{\fnm{Nico} \sur{Pfeifer}}

\author[1,2]{\fnm{Mete} \sur{Akgün}}

\affil[1]{\orgdiv{Medical Data Privacy and Privacy Preserving Machine Learning}, \orgname{University of Tübingen}, \orgaddress{\city{Tübingen}, \postcode{72076}, \country{Germany}}}

\affil[2]{\orgdiv{Institute for Bio-Informatics and Medical Informatics}, \orgname{University of Tübingen}, \orgaddress{\city{Tübingen}, \postcode{72076}, \country{Germany}}}

\abstract{Machine learning models often struggle with generalization in small, heterogeneous datasets due to domain shifts caused by variations in data collection and population differences. This challenge is particularly pronounced in biological data, where data is high-dimensional, small-scale, and decentralized across institutions. While federated domain adaptation methods (FDA) aim to address these challenges, most existing approaches rely on deep learning and focus on classification tasks, making them unsuitable for small-scale, high-dimensional applications. In this work, we propose \textit{freda}, a privacy-preserving federated method for unsupervised domain adaptation in regression tasks. Unlike deep learning-based FDA approaches, \textit{freda} is the first method to enable the federated training of Gaussian Processes to model complex feature relationships while ensuring complete data privacy through randomized encoding and secure aggregation. This allows for effective domain adaptation without direct access to raw data, making it well-suited for applications involving high-dimensional, heterogeneous datasets. We evaluate \textit{freda} on the challenging task of age prediction from DNA methylation data, demonstrating that it achieves performance comparable to the centralized state-of-the-art method while preserving complete data privacy.}


\maketitle

\section{Introduction}\label{Introduction}

Machine learning (ML) has rapidly become a powerful tool with applications across numerous fields, including computational biology and healthcare, where it has shown great potential in solving complex problems \cite{angermueller, greener2022guide, Thumuluri, Valentini}. However, collecting and labeling biological datasets is often challenging, costly, and time-consuming. As a result, many datasets in these fields are small-scale, unlabeled, and heterogeneous, often collected from different sources under varying environmental and experimental conditions—such as different laboratories, hospitals, or institutions \cite{DAreview}. These challenges introduce two critical issues: (1) data from different sources often exhibit distinct statistical distributions while lacking labeled samples, complicating direct model transfer; and (2) privacy regulations and the sensitive nature of biomedical data often restrict data sharing across institutions, necessitating collaborative learning approaches.

Unsupervised Federated Domain Adaptation (FDA) addresses these challenges by collaboratively aligning distributions between training and test data, referred to as source and target domains, respectively, without requiring direct data sharing \cite{farahani2021brief}. The primary motivation for unsupervised FDA is the scarcity of labeled data in the target domain, making it impractical to train models from scratch. Most existing FDA methods aim to mitigate distributional differences between domains \cite{sun2016return, peng2019federated, liu2024ufda, liang2021source}. While deep learning-based FDA methods have achieved success in computer vision \cite{ganin2015unsupervised, NIPS2016_ac627ab1, kd3a, NIPS2016_b59c67bf}, their application to biological data remains limited due to high dimensionality and small sample sizes. Moreover, FDA research has predominantly focused on classification tasks, while regression-based approaches remain significantly underexplored despite their importance in biomedical applications \cite{poplin2018prediction, lundberg2018explainable, li2022graph}.

In this context, we introduce \textit{freda} (\textbf{f}ede\textbf{r}at\textbf{e}d \textbf{d}omain \textbf{a}daptation), a novel method for privacy-preserving, federated unsupervised domain adaptation in regression tasks. Unlike conventional deep learning-based approaches that struggle with data scarcity and high dimensionality, \textit{freda} is the first method to leverage federated training of Gaussian Processes regressors (GPRs), enabling collaborating entities to model complex features without pooling their private data.

Gaussian Processes are particularly well-suited for feature modeling due to their probabilistic nature, providing not only point predictions but also uncertainty estimates in the form of Gaussian-distributed confidence intervals. This property is especially valuable in domain adaptation, where assessing prediction reliability is crucial when transferring knowledge across domains. However, like other kernel-based algorithms, GPRs require pairwise computation of data matrices, making it extremely challenging to train them when data is distributed across entities that cannot share raw samples.

To overcome this, \textit{freda} introduces a novel combination of randomized encoding and secure aggregation, enabling federated training of Gaussian Processes while preserving complete data privacy. By facilitating robust feature modeling without direct access to raw data, \textit{freda} is particularly well-suited for biological datasets, where privacy constraints, limited sample sizes, and data heterogeneity pose significant challenges.

We evaluate \textit{freda} on a challenging benchmark task of age prediction from DNA methylation data. Our results demonstrate that \textit{freda} achieves performance comparable to centralized methods while preserving complete data privacy. By providing a scalable, generalizable, and data-efficient solution for domain adaptation in biological datasets, \textit{freda} enables secure collaboration across institutions. By addressing the challenges of small-scale, heterogeneous, and privacy-sensitive regression problems, our approach significantly expands the applicability of domain adaptation to real-world biomedical applications.

Our contributions are as follows:

\begin{itemize}
\item We propose \textit{freda}, the first method to enable privacy-preserving, federated unsupervised domain adaptation for regression tasks, specifically designed for small-scale, high-dimensional biological datasets.
\item Through a novel combination of randomized encoding and secure aggregation techniques, \textit{freda} is the first method to enable the federated training of GPRs for effective feature modeling while ensuring complete data privacy.
\item We evaluate \textit{freda} on the challenging task of age prediction from DNA methylation data, demonstrating that it effectively models complex feature relationships in small-scale, heterogeneous, and distributed biological datasets, achieving performance comparable to centralized approaches while preserving privacy.
\end{itemize}

\section{Results}\label{results}

To evaluate the performance of our proposed method, we provide a benchmark on the problem of age prediction from DNA methylation data across multiple tissues. This section presents a detailed comparison of \textit{freda} with existing baselines, highlighting its effectiveness in preserving data privacy while achieving competitive predictive accuracy.

\subsection{Overview}

We consider a scenario where the source domain data is distributed across multiple input parties, while the target domain data, held by one party, is unlabeled. The aim is to perform unsupervised domain adaptation on the target domain using labeled source data while preserving data privacy for all parties.

\begin{figure*}[h]
\centering
\includegraphics[width=\textwidth]{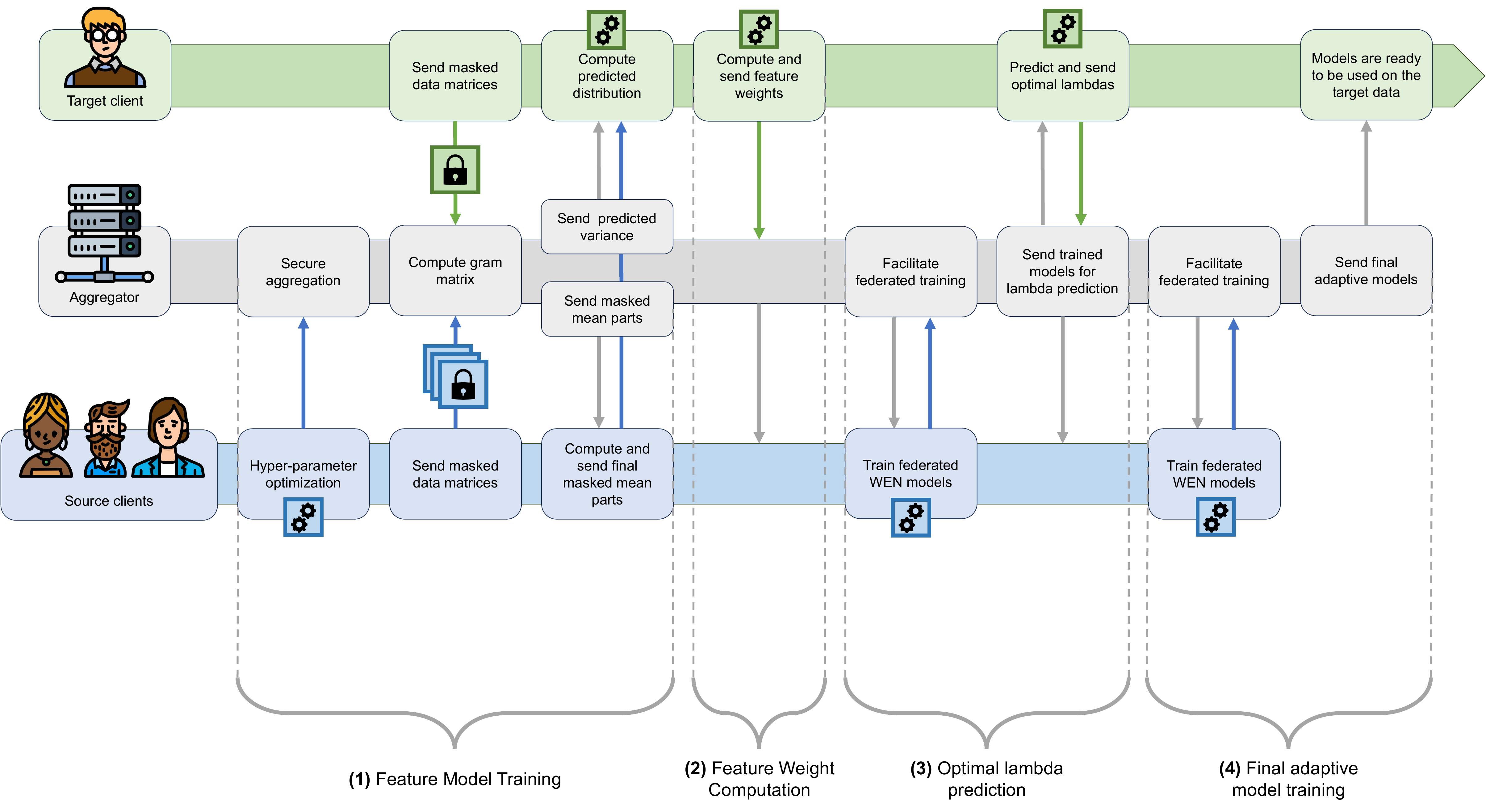}
\caption{Overview of the \textit{freda} framework, in which multiple source domain clients collaborate to perform domain adaptation on the target domain data with the assistance of an aggregator.}
\label{freda_overview}
\end{figure*}

Figure \ref{freda_overview} illustrates the four main phases of our framework:

\begin{enumerate}
    \item \textbf{Feature Model Training: }Each feature in the dataset is modeled separately using federated hyper-parameter optimization and GPRs. In this phase, feature dependencies are learned in a federated and privacy-preserving manner using secure aggregation and randomized encoding.
    \item \textbf{Feature Weight Computation: }The target client calculates feature confidence scores from the predicted distributions of the feature models and transforms them into feature weights. These weights are then shared with the aggregator and distributed to source clients.
    \item \textbf{Optimal Lambda Prediction: }Source clients, with the help of the aggregator, federatively train multiple weighted elastic nets (WEN) with varying regularization parameters (\(\lambda\)). The optimal \(\lambda\) values are identified by the target client and shared with all participants.
    \item \textbf{Final Adaptive Model Training: }Using the optimal \(\lambda\) values, source clients federatively train the final adaptive WEN models, which are then sent to the target client for inference.
\end{enumerate}

This structured approach ensures both privacy preservation and robust performance across distributed data sources.

\subsection{Dataset and Pre-Processing} \label{data}

We utilized DNA methylation data and donor age information from two main sources: the Cancer Genome Atlas (TCGA) \cite{weinstein2013cancer} and the Gene Expression Omnibus (GEO) \cite{edgar2002gene}. For consistency and ease of comparison, we follow the exact preprocessing steps described by Handl et al. \cite{wenda}, including the imputation of missing values, which constituted less than \(0.5\%\) of all samples, as well as dimensionality reduction on the features, reducing the initial set of 466,094 features to 12,980.

We apply a transformation to the chronological ages based on the method proposed by Horvath \cite{horvath2013dna}. For all ages in the training set, we used the function:

\[
F(y) = 
\begin{cases} 
\log(y + 1) - \log(y_{\text{adult}} + 1), & \text{if } y \leq y_{\text{adult}} \\
(y - y_{\text{adult}})/({y_{\text{adult}} + 1}), & \text{otherwise}
\end{cases}
\]

where \( y_{\text{adult}} = 20 \) represents the adult age threshold prior to training. After training, we reversed this transformation using its inverse function, \( F^{-1} \). This transformation is logarithmic for ages below \( y_{\text{adult}} \) and linear for ages above, reflecting that methylation patterns change more rapidly during childhood and adolescence than in adulthood. Finally, we standardized all data to have zero mean and unit variance.

The dataset was then divided into a training (source) set of 1,866 samples and a test (target) set of 1,001 samples. The training set included samples from 19 different tissues, predominantly blood, with donors' ages ranging from 0 to 103 years. The test set initially contained samples from 13 different tissues, including blood and tissues not represented in the training set, such as those from the cerebellum. Following the approach of Handl et al. \cite{wenda}, we aggregated similar tissue types, such as combining \lq blood\rq ,\lq whole blood\rq, and \lq menstrual blood\rq, as well as \lq Brain Medial Frontal Cortex\rq \space and \lq Brain Frontal Cortex\rq, to ensure sufficient sample sizes per tissue type.

\subsection{Baselines}\label{baselines}
We compared the performance of \textit{freda} against the state-of-the-art unsupervised domain adaptation method for biological datasets proposed by Handl et al. \cite{wenda} and a non-adaptive model proposed by Horvath \cite{horvath2013dna}.

\subsubsection{\textit{Wenda} Baseline}  

\textit{Wenda} (weighted elastic net for domain adaptation) is a state-of-the-art method for unsupervised domain adaptation on small-scale high-dimensional biological datasets \cite{wenda}. Like our approach, it leverages the dependency structure between inputs across source and target domains, penalizing discrepant features while emphasizing robust ones. Despite its strong performance over non-adaptive models, \textit{wenda} assumes simultaneous access to both domains, hence it is only suitable for centralized settings.

\textit{Wenda} has three key parameters: the weighting parameter \(k\), the elastic net mixing parameter \(\alpha\), and the regularization parameter \(\lambda\). Following \cite{wenda}, we fix \(\alpha = 0.8\), while \(\lambda\) is computed using prior knowledge on tissue similarity (\textit{wenda-pn}), as cross-validation on the target domain is infeasible in an unsupervised setting. For \(k\), we select \(k = 3\) based on both our experiments and prior work \cite{wenda}.  

\subsubsection{Non-Adaptive Baseline}

For our non-adaptive baseline, we adopt the method proposed by Horvath \cite{horvath2013dna}, which combines the elastic net with a least-squares fit. The idea is to first fit a standard elastic net and then apply a linear least-squares fit based only on features that obtained non-zero coefficients in the elastic net. This baseline was first proposed by Horvath \cite{horvath2013dna} for age prediction from DNA methylation data, where he demonstrated that using an elastic net followed by a least-squares fit resulted in improved performance on his dataset. We refer to this non-adaptive method as \textit{en-ls}.

\subsection{Setup for \textit{Freda}}  

We consider a distributed setting with multiple source domain clients, a target client, and an aggregator, which has no data. The labeled source domain data is distributed across 2, 4, or 8 source clients, and we evaluate \textit{freda} in each of these settings.  

\subsubsection{Data Distribution}  

Source domain data is assigned uniformly at random among the source clients. Given that the DNA methylation dataset contains 1,866 training samples, each client receives approximately 933, 466, or 233 samples in the 2, 4, and 8-client settings, respectively.  

To assess the robustness of \textit{freda}, we do not consider tissue types when distributing data. Due to the inherent imbalance of DNA methylation data across tissues (Section~\ref{data}), this results in some clients having only a few or no samples from certain tissues.  

\subsubsection{Setup for Weighted Elastic Net Models}  

The final weighted elastic net model is trained for 100 global iterations. In each iteration, source clients update their local model for 20 epochs before the aggregator securely updates the global model.  

Furthermore, we apply an exponential learning rate decay based on global iterations \cite{yan2022seizing}:

\begin{equation}  
\eta(t) = \eta_0 \left( \frac{\eta_f}{\eta_0} \right)^{\frac{t}{T}}  
\end{equation}  

where $\eta(t)$ is the learning rate at iteration $t$, $T$ is the total number of iterations, and $t$ is the current iteration. We set the initial and final learning rates to $\eta_0 = 1 \times 10^{-4}$ and $\eta_f = 1 \times 10^{-5}$, respectively.  

\subsubsection{Parameter Selection in \textit{Freda}} \label{external_parameters}  

\textit{Freda} has three key parameters: the weighting parameter \(k\), the elastic net mixing parameter \(\alpha\), and the regularization parameter \(\lambda\). We fix \(\alpha = 0.8\) as a design choice and determine \(\lambda\) using the prior knowledge approach (Section~\ref{pn}) proposed by Handl et al. \cite{wenda}.  

For tissue similarity calculations, we use data from the GTEx consortium, which provides genotype and gene expression data across 42 human tissues \cite{gtex2017genetic}, following the methodology in \cite{wenda}. In the federated setting, only the target domain owner requires access to tissue similarity information.  

To evaluate performance, we follow the evaluation strategy of Handl et al. for \textit{wenda-pn} \cite{wenda}, iteratively splitting test tissues into subsets: one for fitting domain similarity relationships and another for evaluation (Section~\ref{pn}). We iterate over all three-tissue combinations with at least 20 training samples, assessing performance on the remaining tissues. Based on our experiments, the optimal weighting parameter is \(k = 3\).

\subsection{Experiments} \label{experiments}

\begin{figure}[h]
\centering
\includegraphics[width=0.95\textwidth]{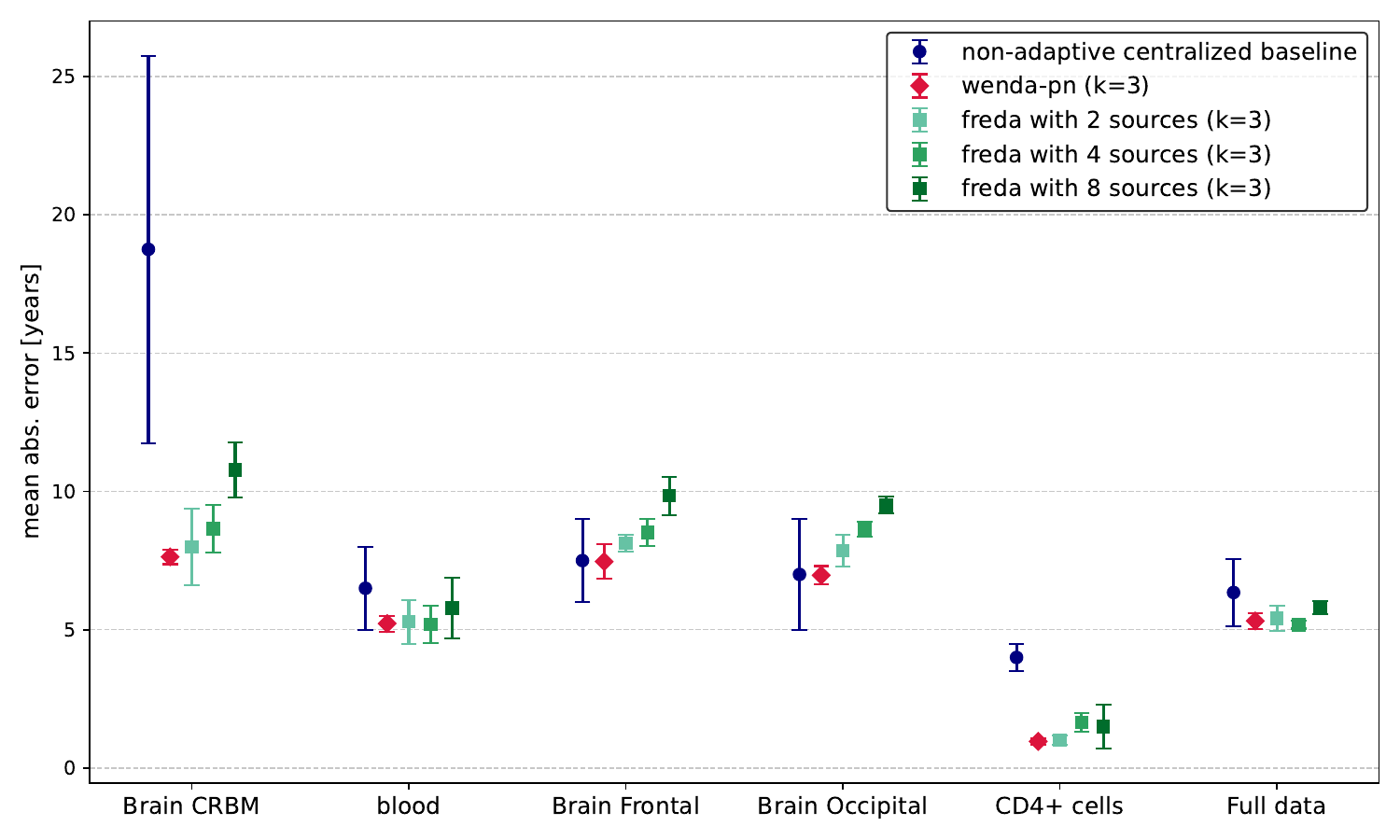}
\caption{Mean absolute error per target tissue, as well as on full target data for the non-adaptive and non-federated baseline \textit{en-ls}, \textit{wenda-pn} with \(k=3\), and \textit{freda} with \(k=3\), across 2, 4, and 8 source parties.}
\label{all_comparison}
\end{figure}

We compare the performance of \textit{freda} against \textit{wenda-pn} and the non-private, non-adaptive baseline model \textit{en-ls}, as described in Section \ref{baselines}. The main performance metric is the Mean Absolute Error (MAE) of the predicted chronological ages of the tissues. For \textit{wenda-pn}, we calculate the MAE only on samples not used for fitting the tissue similarity-\(\lambda\) relationship, reporting the mean and standard deviation across all splits. Similarly, for \textit{freda}, we report the MAE exclusively for the target client's tissues that were not part of the similarity-\(\lambda\) fit, along with the mean and standard deviation over all splits.

For the non-adaptive baseline \textit{en-ls}, Handl et al. emphasize that the heterogeneous nature of the data and the random splitting of the training data used for 10-fold cross-validation significantly influence its performance. Therefore, we follow their approach and report the mean \(\pm\) standard deviation over 10 runs for \textit{en-ls}. For \textit{wenda-pn}, the mean \(\pm\) standard deviation is calculated over all splits of the test tissues where the tissue of interest was included in the evaluation set. For \textit{freda}, we report the mean \(\pm\) standard deviation for each setting (2, 4, and 8 sources) over 5 different uniform random distributions of source data across the source parties, considering all splits where the tissue of interest was included in the evaluation set.

For \textit{wenda-pn}, Handl et al. \cite{wenda} treat each tissue in the test dataset as a separate target domain, training the final weighted elastic net models independently for each tissue. Specifically, Handl et al. \cite{wenda} compute the average confidences, as defined in Equation \ref{confidences}, only over the samples of the same tissue and train a separate model for each tissue, always using the entirety of the training (source) data but applying tissue-specific feature weights. We follow the same approach for \textit{freda} in all our experiments, where the clients inside the federated learning system train a separate weighted elastic net model for each tissue in the target domain (for further information see Section \ref{freda}).


The performance of \textit{en-ls}, \textit{wenda-pn}, and \textit{freda} for 2, 4, and 8 source parties on the relevant tissues of the target domain, as well as on all samples of the target domain data, is shown in Figure \ref{all_comparison}. For the full target dataset, the centralized baseline methods \textit{en-ls} and \textit{wenda-pn} yield an MAE of \(6.34 \pm 1.21\) and \(5.31 \pm 0.29\), respectively. These results indicate that when the entire target domain data is considered, \textit{wenda-pn} provides only a slight improvement in performance compared to the non-adaptive \textit{en-ls}. The effect of distribution shift is most visible when we observe the performance of our baselines on cerebellum samples. As shown in Figure \ref{all_comparison}, the non-adaptive \textit{en-ls} yields a significantly higher MAE on cerebellum samples compared to other tissues.

\begin{figure*}[h]
\centering
\includegraphics[width=0.328\textwidth]{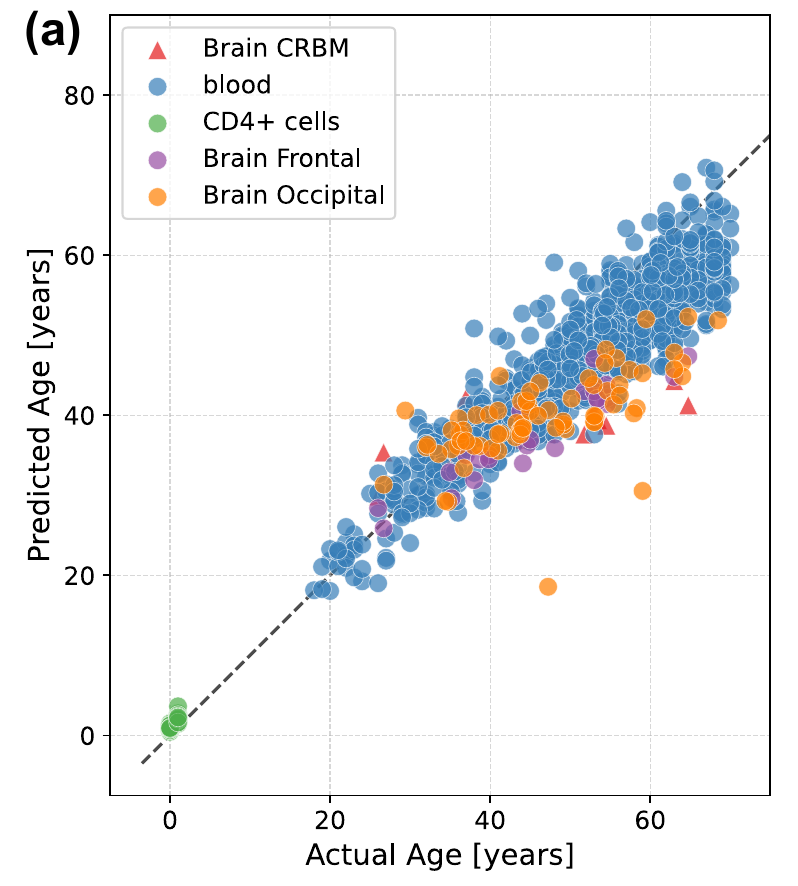}
\includegraphics[width=0.328\textwidth]{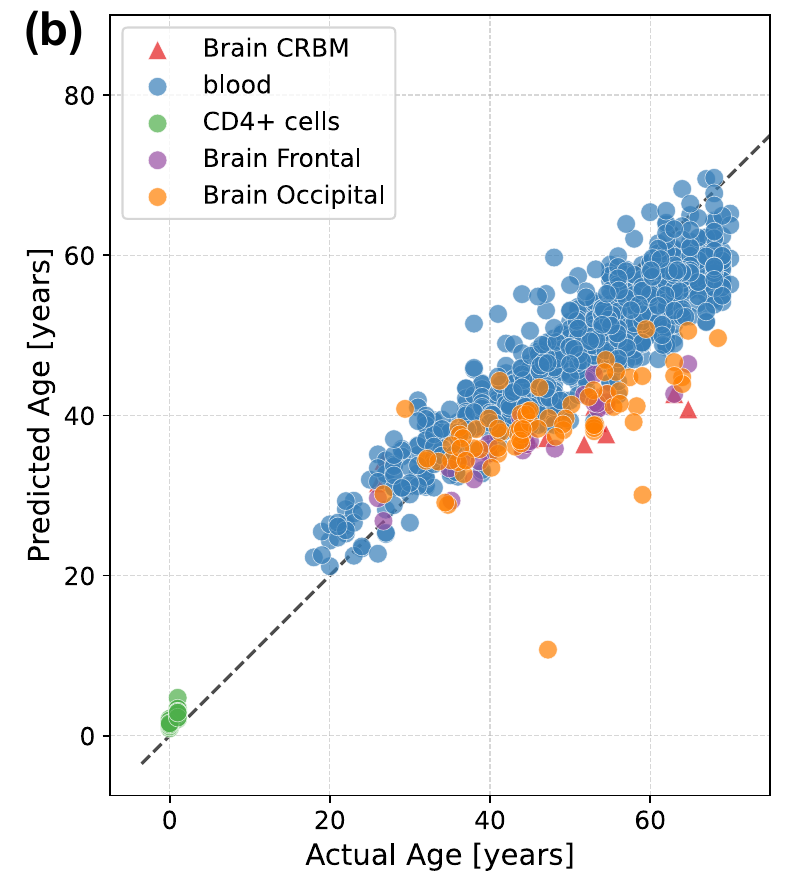}
\includegraphics[width=0.328\textwidth]{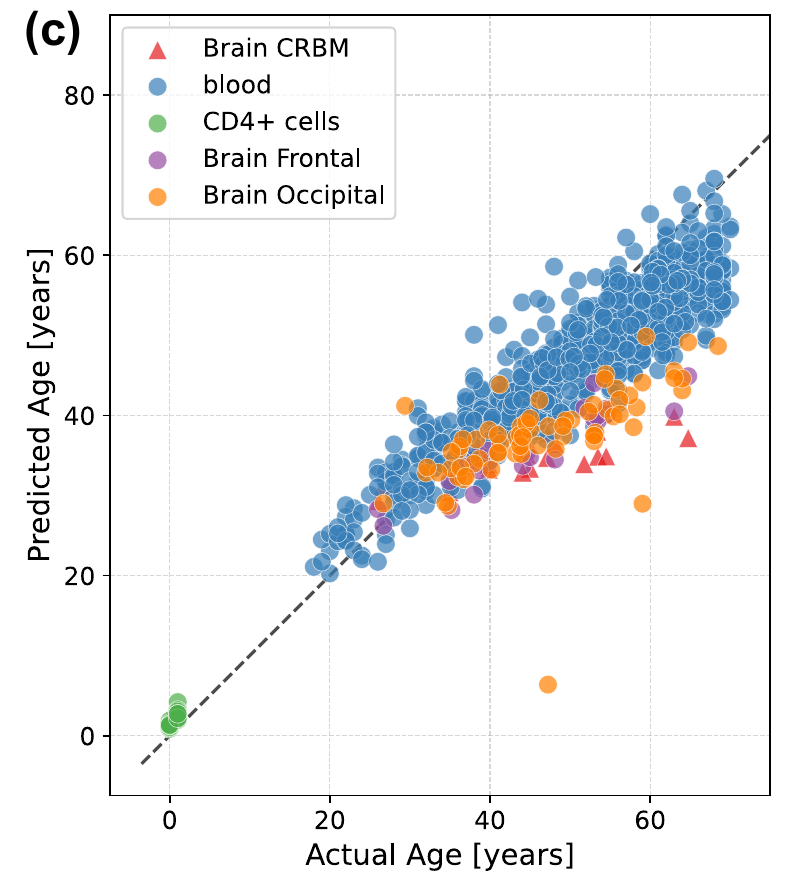}
\includegraphics[width=0.328\textwidth]{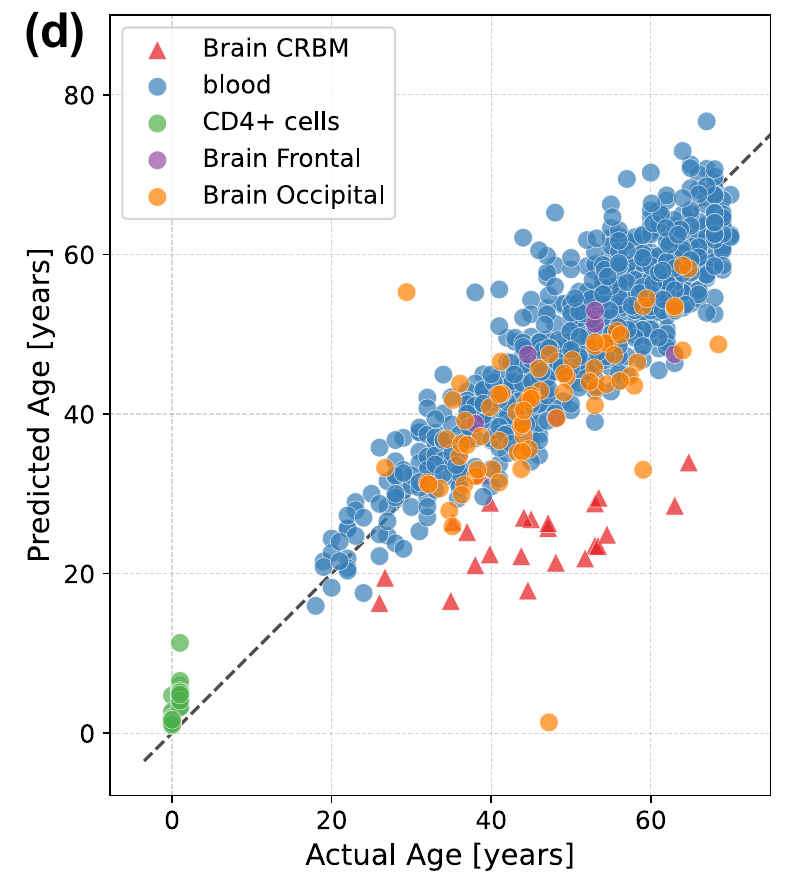}
\includegraphics[width=0.328\textwidth]{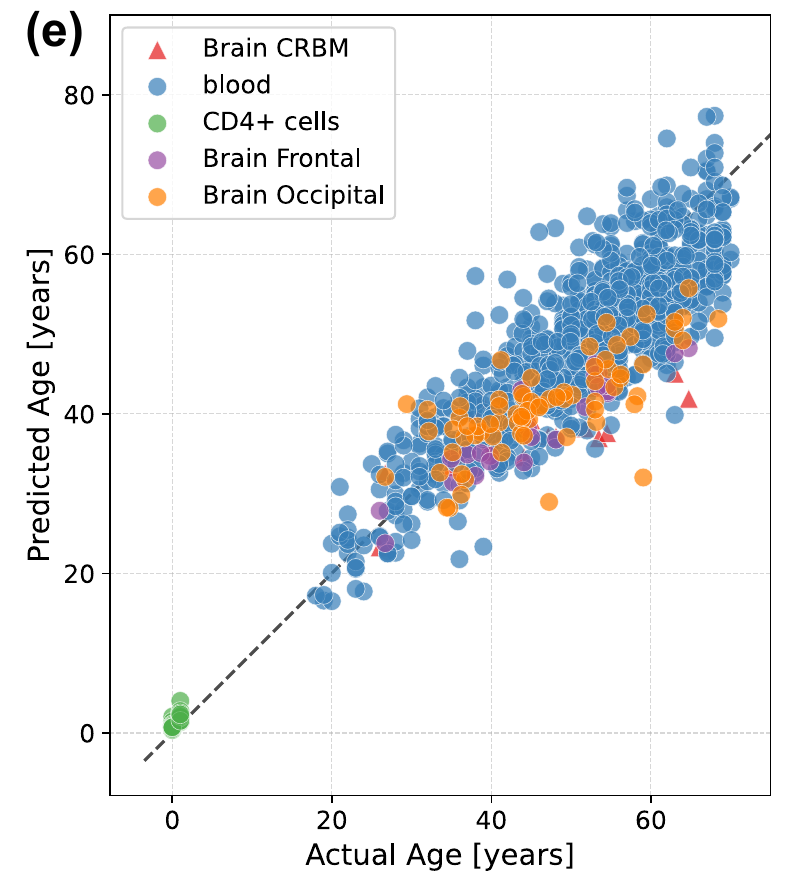}

\caption{Predicted versus true chronological age under various settings. Figures \textbf{(a)}, \textbf{(b)}, and \textbf{(c)} correspond to \textit{freda} with \(k=3\) for 2, 4, and 8 source parties, respectively. Predictions are averaged over all splits where the tissue of interest was included in the evaluation set, as well as over 5 different distributions for each setting. Panels \textbf{(d)} and \textbf{(e)} correspond to \textit{en-ls} and \textit{wenda-pn}, respectively. For \textit{en-ls}, predictions are averaged over 10 runs of 10-fold cross-validation, while for \textit{wenda-pn}, predictions are averaged over all splits where the tissue of interest was included in the evaluation set.}

\label{freda_preds}
\end{figure*}

Figure \ref{freda_preds} show the predicted versus true ages for the samples of the target domain data, colored by tissue, for \textit{freda} with \(k=3\) for 2, 4, and 8 source parties, \textit{en-ls} and \textit{wenda-pn}, respectively. From Figures \ref{freda_preds}d and \ref{freda_preds}e, we can clearly see that both centralized methods perform well on most tissues, except for \textit{en-ls} on cerebellum samples. As shown in Figure \ref{freda_preds}d, the ages predicted by \textit{en-ls} for cerebellum samples are consistently lower than the true chronological ages. In contrast, Figure \ref{freda_preds}e demonstrates that \textit{wenda-pn} achieves much closer alignment between the predicted and true ages for cerebellum samples.

Additionally, for the remaining target domain tissues, the predictions of \textit{wenda-pn} are comparable to those of \textit{en-ls}, as confirmed by the quantitative results in Figure \ref{all_comparison}. \textit{Wenda-pn} not only yields significantly lower errors than \textit{en-ls} on cerebellum samples but also maintains similar or better performance on other test tissues. Specifically, on cerebellum samples, \textit{en-ls} produces a mean absolute error (MAE) of \(7.63 \pm 0.26\). These results highlight the significance of improving prediction performance on cerebellum samples without a drop in performance on other tissues.

Our experimental results, presented in Figures \ref{all_comparison} and \ref{freda_preds}, are consistent with the findings reported by Handl et al. \cite{wenda}. Handl et al. highlight the difficulty of predicting the age of cerebellum samples, noting that these samples are not represented in the training data and are known to be biologically distinct, even from other brain tissues, in terms of function and gene expression patterns \cite{fraser2005aging, gtex2017genetic}. Hence, our evaluation focuses on whether federated privacy-preserving domain adaptation, as implemented by \textit{freda}, can achieve comparable performance on these samples to the centralized method \textit{wenda-pn}.

For the full target dataset, \textit{freda} achieves a MAE of \(5.41 \pm 0.44\), \(5.41 \pm 0.44\), and \(5.81 \pm 0.24\) for the 2, 4, and 8 source domain settings, respectively. These results indicate that, when considering the full target domain data, \textit{freda} provides a performance level almost identical to that of \textit{wenda-pn} and consistently better than \textit{en-ls} across all configurations, despite operating in a distributed environment.

Focusing on the primary metric for evaluating the success of unsupervised domain adaptation, the performance of \textit{freda} on cerebellum samples compared to our centralized baselines is shown in Figure \ref{all_comparison}. Despite running in a distributed setting, \textit{freda} achieves comparable performance to \textit{wenda-pn} on cerebellum samples for the 2 and 4 source domain scenarios, yielding an MAE of \(7.99 \pm 1.39\) and \(8.64 \pm 0.86\), respectively. In the 8 source domain scenario, \textit{freda} yields an MAE of \(10.77 \pm 0.99\), which is only slightly worse than that of \textit{wenda-pn}. However, even when the source domain data is distributed across 8 parties, \textit{freda} still significantly outperforms the non-adaptive centralized baseline \textit{en-ls}. This observation is further supported by the prediction results shown in Figure \ref{freda_preds}. Figures \ref{freda_preds}a, \ref{freda_preds}b, and \ref{freda_preds}c shows the predicted versus true ages for the target domain data samples, colored by tissue, for \textit{freda} under the 2, 4, and 8 source party settings, respectively. Comparing the predictive performance of \textit{freda} with our baselines, it is clear that \textit{freda} performs well across all tissues. Moreover, the consistently low predicted ages for cerebellum samples relative to their true chronological ages, as observed with \textit{en-ls}, are not present in any of the \textit{freda} settings. In comparison with \textit{wenda-pn}, the results in Figure \ref{freda_preds} show that \textit{freda} performs similarly to \textit{wenda-pn}, while effectively addressing the distribution shift associated with cerebellum samples, even in a distributed setting where the source domain data is spread across up to $8$ different parties.

\section{Discussion}\label{discussion}  

In this study, we introduce \textit{freda}, the first method to enable privacy-preserving, federated unsupervised domain adaptation for regression tasks in high-dimensional, small-scale biological datasets.

Existing FDA methods predominantly rely on deep learning models, which require large datasets and often struggle with high-dimensional, small-sample-size biological data. Furthermore, these methods primarily focus on classification tasks, leaving regression-based FDA approaches largely underexplored. In contrast, \textit{freda} introduces a novel combination of randomized encoding and secure aggregation to enable the first privacy-preserving, federated training of Gaussian Processes. By leveraging Gaussian Processes, \textit{freda} effectively models complex feature relationships in high-dimensional small-scale datasets while preserving complete data privacy, making it particularly well-suited for biomedical applications.

We evaluate \textit{freda} on a challenging benchmark task of age prediction from DNA methylation data. Our experimental results demonstrate that \textit{freda} performs comparably to centralized state-of-the-art domain adaptation method for small-scale and high-dimensional datasets while significantly outperforming the non-adaptive centralized baseline, even when source domain data is distributed across multiple parties. Notably, for cerebellum samples, which exhibit a challenging distribution shift, \textit{freda} effectively models complex feature relationships and adapts to the shift without requiring raw data sharing. This highlights its potential for real-world applications where data heterogeneity and privacy constraints pose significant challenges.  

A key contribution of our work is showing that privacy-preserving federated training of Gaussian Processes is not only feasible but also highly effective for distributed domain adaptation in biological datasets. Traditional Gaussian Process models, while well-suited for feature modeling, have never been scaled to distributed settings due to their reliance on pairwise data computations. \textit{Freda} is the first method to overcome this limitation by enabling secure and collaborative training, ensuring that institutions can benefit from improved feature modeling without compromising data privacy.

\section{Related Work}\label{wenda}

\subsection{Domain Adaptation}

Domain adaptation, particularly in the unsupervised setting, has been extensively studied in recent years \cite{yue2023make,wang2018deep, weng2023open}. The goal of domain adaptation is to transfer knowledge from a source domain to a target domain, where labeled data in the target domain is scarce or unavailable \cite{chen2023multi}. While numerous methods exist, the majority of these focus on image classification problems, where the source data is typically abundant and the number of features is relatively small.

Notable unsupervised domain adaptation techniques often aim to align feature representations across domains. One common approach involves adversarial training to minimize the domain discrepancy. For instance, the Domain-Adversarial Neural Network proposed by Ganin et al. \cite{ganin2016domain} uses a gradient reversal layer to align source and target feature distributions via adversarial learning, significantly improving classification performance on tasks like digit recognition. Another popular method is Maximum Mean Discrepancy based alignment, proposed by Long et al.'s Deep Adaptation Networks, which align the distributions of deep features between domains using kernel-based metrics \cite{long2015learning}. Additionally, self-supervised learning approaches such as CDAN \cite{long2018conditional} integrate task-specific predictions into domain alignment to capture conditional dependencies. These methods are widely adopted in benchmarks like Office-31, VisDA, or digit datasets, showcasing high performance in handling distribution shift in image classification.

\subsection{Domain Adaptation for Regression}

Unlike classification problems, domain adaptation for regression has received less attention over the years, primarily due to the challenges in aligning continuous output spaces. The recently proposed Distribution-Informed Neural Network (DINO) \cite{wu2022distribution} framework introduces a unified approach to tackle this issue by leveraging distribution-aware neural networks. Empirical results demonstrate that DINO achieves state-of-the-art performance across domain adaptation regression benchmarks, highlighting its effectiveness in mitigating distribution shifts in regression tasks \cite{wu2022distribution}. Still, DINO has been tested on image datasets with high sample sizes and low feature counts, while effective in such scenarios, these datasets are not representative of the high-dimensional, low-sample-size challenges faced in biological datasets.  

A key state-of-the-art method for unsupervised domain adaptation in high-dimensional, small-sample datasets is \textit{wenda} \cite{wenda}. Unlike deep learning-based approaches that require large datasets, \textit{wenda} is specifically designed for tabular biological data and operates under a relaxed covariate shift assumption, allowing for domain adaptation even when certain features behave differently across domains. It achieves this by estimating feature dependencies in the source domain, assessing their consistency in the target domain, and applying adaptive regularization to account for domain shifts. In their study, Handl et al. demonstrated that \textit{wenda} significantly improves adaptation performance in DNA methylation-based age prediction, particularly for cerebellum samples, which exhibit substantial distributional shifts. Due to its effectiveness in this domain, we use \textit{wenda} as a baseline method to compare with \textit{freda}.

\subsection{Federated Domain Adaptation}

Federated domain adaptation extends the principles of domain adaptation to the FL setting, where data is located on decentralized devices or institutions, and privacy constraints prevent data sharing. In this context, methods must tackle both domain shifts, the heterogeneity inherent in FL environments, and the privacy concerns. Multiple federated domain adaptation techniques have been proposed in the recent years. For instance, PartialFed \cite{sun2021partialfed} introduces a strategy where only a subset of global model parameters is loaded on local clients, allowing for dynamic mixing of global and client-specific parameters. This approach enhances performance on cross-domain classification and detection tasks, demonstrating significant improvements on image datasets such as Office-Home and UODB \cite{sun2021partialfed}.

Similarly, FedGP \cite{jiang2024principled} proposes a gradient-projection-based aggregation method to enhance the performance of the model on a target client with limited data. By filtering out noisy source gradients and optimally combining target and source gradients, FedGP achieves notable performance improvements under both domain shift and data scarcity scenarios. Experiments on real-world image datasets verify its robustness and efficacy in FDA settings \cite{jiang2024principled}.

Still, these studies primarily focus on image datasets with abundant samples and relatively low-dimensional features, which may not translate effectively to domains such as computational biology, where data is high-dimensional and samples are limited.

\section{The \textsl{Freda} Method}\label{freda}

We consider the following distributed setting: there are \(N\) source domain clients, each with a local labeled dataset \(X^{s_i} = \{(x^{s_i}_j, y^{s_i}_j)\}_{j=1}^{n_i}\) and a sample size of \(n_i\). The entire source domain data, distributed across the \(N\) clients, is denoted as \(X^S = \bigcup_{i=1}^N X^{s_i}\). Similarly, there is a target client with a dataset \(X^T = \{x^t_m\}_{m=1}^{n_t}\) containing \(n_t\) samples for the same prediction task, but without any available labels. In both the source and target domain datasets, the samples \(\{x^{s_i}_j\}\) and \(\{x^t_m\}\) are \(\mathcal{P}\)-dimensional vectors, where \(\mathcal{P} \in \mathbb{N}\), and the labels \(\{y^{s_i}_j\}\) are scalars.

The goal of our framework is to leverage the inputs and labels from the source domains, along with the inputs from the target domain, to develop a model that achieves high performance on the target domain but without requiring explicit data sharing between parties.

\textit{Freda} operates with the same underlying principles as \textit{wenda}. The process begins with collaboratively learning the input dependency structure in a distributed setting by training feature models. Next, confidence scores are computed from the predictions of these feature models, which are then transformed into feature weights. Subsequently, optimal lambda prediction is performed by federatively training multiple models on the source domain using varying regularization parameters. This step enables the target domain owner to determine the optimal regularization parameter. Finally, the method concludes with the federated training of the final adaptive model on the source domains.

For simplicity, we describe our method using a single domain in the target data; however, similar to the approach of Handl et al. \cite{wenda}, \textit{Freda} can be extended to multiple target domains inside the target data, as demonstrated in Section \ref{results}.

\subsection{Feature Model Training} \label{feature_models}

The performance of our framework relies on capturing the input dependency structure between the source and target domain data in a distributed setting while maintaining data privacy. To achieve this, we leverage Bayesian models, specifically Gaussian Process Regressors, to model the dependency structure. For each feature \(f\), we train a GPR model \(g_f\) that predicts \(f\) based on all other features, using the source domain inputs from all source clients as training data.

The training data for a given feature \(f\) includes \(X^{S}_{\neg f}\), the entire source domain data with the column corresponding to feature \(f\) removed, as inputs, and \(X^{S}_f\), the corresponding feature vector, as labels. For new data points \(X^T_{\neg f} = [x^t_{1, \neg f}, \ldots, x^t_{n_t, \neg f}]\), the target domain data with the column for feature \(f\) removed, the goal is to predict the corresponding feature vector \(X^t_f = [x^t_{1, f}, \ldots, x^t_{n_t, f}]\).

For a specific source domain \(i\), the vector containing feature \(f\) is denoted \(X^{s_i}_f = [x^{s_i}_{1, f}, \ldots, x^{s_i}_{n_i, f}]\), while the \((n_i \times (\mathcal{P} - 1))\)-matrix of remaining features is \(X^{s_i}_{\neg f} = [x^{s_i}_{1, \neg f}, \ldots, x^{s_i}_{n_i, \neg f}]\). Thus, for a given feature \(f\), the GPR model \(g_f\) provides a closed-form predictive distribution:

\begin{equation}
     \mathcal{N}(K_* K^{-1} y, K_{**} - K_* K^{-1} K_*^\top)
\label{gpr}
\end{equation}

where:

\begin{align}
K      &= k(X^{S}_{\neg f}, X^{S}_{\neg f}) + \sigma_n^2 \mathbbm{1}_{n_S} \nonumber \\
K_*    &= k(X^{S}_{\neg f}, X^T_{\neg f}) \label{Ks} \\
K_{**} &= k(X^T_{\neg f}, X^T_{\neg f}) \nonumber \\
y &= X^{S}_f \nonumber
\end{align}

Here, \(k(.,.)\) computes the linear kernel with the variance of the prior on the coefficients \(\sigma_p^2\) between the given input matrices \cite{williams2006gaussian}. Additionally, \(n_S\) denotes the total number of samples in the entire source domain dataset \(X^S\). Unlike traditional supervised regression models that predict a single value for a given input, GPRs provide a full predictive distribution as output \cite{seeger2004gaussian}, which we later use to compute feature weights. This GPR model involves two hyper-parameters that must be optimized for the best performance: the variance of the kernel, \(\sigma_p^2\), and the variance of the additive Gaussian noise, \(\sigma_n^2\), from the closed-form solution in Eq. \ref{gpr}. The optimal values of these hyper-parameters are determined by maximizing the marginal likelihood for each feature. For a specific source client \(i\), and the covariance matrix \(K=k(X^{s_i}_{\neg f}, X^{s_i}_{\neg f}) + \sigma_n^2 \mathbbm{1}_{n_i}\) from Eq. \ref{Ks}, source client \(i\) maximizes:

\begin{equation}
    \log \mathcal{L}(X^{s_i}_{f}|X^{s_i}_{\neg f}) = -\frac{1}{2} (X^{s_i}_{f})^\top K^{-1} X^{s_i}_{f} - \frac{1}{2} \log |K| - \frac{n_i}{2} \log(2\pi)
    \label{hyper-params}
\end{equation}

Training feature models is straightforward when both target and source domains are accessible simultaneously. However, significant challenges arise when these datasets are distributed.

The first challenge is that, if the source domain is distributed across multiple entities, the optimization of hyper-parameters shown in Eq. \ref{hyper-params} cannot be performed across the entire source domain. The second, and more complex, challenge is that due to the distribution of the source and target domains, the closed-form solution of the GPR model (as shown in Eq. \ref{gpr}) cannot be computed directly. Since GPRs are non-parametric machine learning algorithms, obtaining predictions requires computing the three matrices in Eq. \ref{Ks}, namely \(K\), \(K_*\), and \(K_{**}\).

In our setting, computing \(K_{**}\) is straightforward and can be performed locally by the owner of the target domain, as it requires only the target domain data. However, computing \(K\) and \(K_*\) as well as the predictive mean of the feature model \(K_* K^{-1} y\) presents significant challenges. Computing \(K\) is challenging because, although it only requires source domain data, the data is distributed across multiple entities, and the entire matrix product of \(K\) cannot be computed trivially. Instead, it must be computed collaboratively among all source domain owners while preserving privacy. Similarly, computing \(K_*\) is challenging because it requires access to both source domain data and target domain data. Lastly, computing \(K_* K^{-1} y\) is not straightforward because the feature column for the entire source domain data, \(y=X^{S}_f\), is distributed between \(N\) source clients. Moreover, explicitly sharing \(y\) would compromise privacy since the aggregator could reconstruct the data of all source clients after the feature model training phase is completed, as each feature is modeled independently.

\textit{Freda} addresses all of these challenges by employing secure aggregation and a special masking scheme for matrix product computation called FLAKE \cite{hannemann2023privacy}.

\subsubsection{Federated Hyper-Parameter Optimization} \label{hp_opt}

To optimize the hyper-parameters required for the GPR models in a distributed setting, specifically the variance of the prior on the coefficients \(\sigma_p^2\) in the kernel function and the variance of the additive noise \(\sigma_n^2\) in the closed-form solution (Eq. \ref{gpr}), we employ secure aggregation \cite{bonawitz2016practical}.

Secure aggregation refers to a range of techniques designed to compute a multiparty sum without exposing any individual participant's data in plain text, even to the aggregator \cite{bonawitz2016practical}. While advanced privacy-enhancing technologies like Fully Homomorphic Encryption (FHE) or Secure Multi-party Computation (MPC) can achieve secure aggregation, \textit{Freda} adopts a more practical approach using a method known as zero-sum masking.

Zero-sum masking, as proposed by Bonawitz et al. \cite{bonawitz2016practical}, allows participants to securely compute the sum of their inputs while ensuring the privacy of each party’s data. In this approach, each participant masks its data with a random value, making the masked values indistinguishable from random noise. The random values are chosen such that, when the contributions from all participants are summed, the masking terms cancel out, revealing only the aggregated sum of the original data.

The process is as follows: for each party \(i\), a random mask \(r_i\) is generated and added to the data \(d_i\), resulting in \(d_i + r_i\). Each party shares its masked data with the aggregator while simultaneously exchanging the necessary random masks \(r_i\) with other parties. When the aggregator computes the sum of all masked values, the random masks cancel out across participants, revealing only the aggregated sum \(d_1 + d_2 + \cdots + d_N\). This method ensures data privacy without requiring complex cryptographic operations, making it efficient and suitable for large-scale federated learning systems.

In \textit{Freda}, the optimization of the hyper-parameters \(\sigma_p^2\) and \(\sigma_n^2\) is performed locally by each source client. Each source client \(i\) maximizes the marginal likelihood for every feature in its local dataset \(\{x^{s_i}_j\}_{j=1}^{n_i}\), following Eq. \ref{hyper-params}, to obtain \(\sigma_{p,i}^2\) and \(\sigma_{n,i}^2\). These locally optimized hyper-parameter pairs are then aggregated using zero-sum masking, enabling the aggregator to compute global averages for \(\sigma_p^2\) and \(\sigma_n^2\) as if the marginal likelihood had been maximized over the entire source domain data.

\subsubsection{Federated GPR Computation} \label{gpr_computation}

In the GPR model prediction process, the most challenging components to compute are \(K\), \(K_*\), and their product with the global feature vector \(y=X^S_f\), as these require access to both the source domain data and the target domain data to calculate the matrix product. A naive plaintext approach would compromise privacy. To overcome this, we utilize FLAKE \cite{hannemann2023privacy}, a framework for secure and private matrix product computation, which allows us to calculate the product of matrices from the source and target domains without disclosing their plaintext values.

\paragraph{Privacy-Preserving Masking Process}

We use special masking matrices to hide the input matrices of the matrix product and reveal only the result of this multiplication to the aggregator. This protects the privacy of the input matrices as well as their dimensionality. Specifically, both the source clients and the target client share a common seed to generate a shared mask matrix \(M \in \mathbb{R}^{d \times \mathcal{P}}\), where \(d\) is a higher-dimensional space than the original feature space. Each client \(p\) locally computes a left inverse \(L_p \in \mathbb{R}^{\mathcal{P} \times d}\) for \(M\), such that \(L_p M = I_d\), and applies the mask to its data:

\[
x_p' = x_p L_p (M M^\top)^{1/2},
\]

where \(x_p\) represents the local dataset of client \(p\) (\(x_p = X^{s_i}\) for a source client or \(x_p = X^T\) for the target client). The masked data \(x_p'\) is sent to the aggregator.

\paragraph{Secure Gram Matrix Computation}

The aggregator computes the Gram matrix for all clients using the masked data. For a pair of clients \(p\) and \(q\), the Gram matrix is computed as:

\begin{equation}
\begin{split}
    x_p x_q^\top &= x_p^{'} (x_q^{'})^\top \\
    &= (x_p L_p (M M^\top)^{1/2}) (x_q L_q (M M^\top)^{1/2})^\top \\
    &= x_p L_p (M M^\top)^{1/2} ((M M^\top)^{1/2})^\top L_q^\top x_q^\top \\
    &= x_p (L_p M) (M^\top L_q^\top) x_q^\top \\
    &= x_p x_q^\top,
\end{split}
\end{equation}
where the masking terms cancel out, revealing only the product \(x_p x_q^\top\) without exposing the individual data of the clients.

Using the computed Gram matrix \(G_{pq}\), the aggregator calculates \(K\) and \(K_*\) as follows:
\begin{align*}
    K &= \sigma_p^2 G_{pq} + \sigma_n^2 \mathbbm{1}_{n_S}, \\
    K_* &= \sigma_p^2 G_{pq}.
\end{align*}

\paragraph{Computing the Predicted Mean}

Computing the predicted mean \(K_* K^{-1} y\) is challenging as the aggregator does not have access to the label vector \(y\) (feature column for the modeled feature \(g_f\)), which remains distributed across source clients. Explicitly sharing \(y\) would also compromise privacy since the aggregator could reconstruct the data of all source clients after the feature model training phase is completed, as each feature is modeled independently. To address this, \textit{Freda} employs randomized encoding.

\begin{enumerate}
    \item \textbf{Aggregator Step:} The aggregator computes the masked intermediate result \(C K_* K^{-1}\), where \(C \in \mathbb{R}^{n_t \times n_t}\) is a random mask matrix, and sends \(C^{-1}\) to the target client.
    \item \textbf{Source Client Collaboration:} The masked matrix \(C K_* K^{-1}\) is split into sub-matrices corresponding to each source client’s data represented as \( (C K_* K^{-1})^{s_i}\). Each source client \(i\) computes:
    \[
    (C K_* K^{-1})^{s_i} X^{s_i}_{f},
    \]
    This computation results in a \(n_t \times 1\) vector, which each source client sends to the target client. 
    \item \textbf{Target Client Aggregation:} The target client aggregates the received vectors and removes the mask using \(C^{-1}\):
    \[
    K_* K^{-1} X^{S}_{f} = C^{-1} \sum_{i=1}^N (C K_* K^{-1})^{s_i} X^{s_i}_{f}.
    \]
\end{enumerate}

This process enables the target client to compute the predicted mean while ensuring that the privacy of individual source clients' data is preserved. Additionally, it protects against source clients inferring any information from the sub-matrices of the intermediate matrix product \(K_* K^{-1}\).

\paragraph{Computing the Predicted Variance}

The predicted variance \(K_{**} - K_* K^{-1} K_*^\top\) is computed using the Gram matrix \(G_{pq}\) and sent to the target client by the aggregator. With both the predicted mean and variance, the target client obtains the complete closed-form solution of the GPR model.

\subsection{Feature Weight Computation} \label{confidence_score_computation}

This step is performed entirely by the target client. After obtaining the predictions from the feature models, the target client computes the confidence score for each feature based on the distribution predicted by the feature models. These confidence scores are then used to determine the weight of each feature.

For a sample from the target domain, denoted as \(\tilde{x}_m\), and a given feature \(f\), let \(\tilde{x}_{m,f}\) represent the value of feature \(f\) in \(\tilde{x}_m\), and \(\tilde{x}_{m,\neg f}\) represent the values of all other features in \(\tilde{x}_m\). Given \(\tilde{x}_{m,\neg f}\), the feature model \(g_f\) outputs a posterior distribution that describes the expected value of \(\tilde{x}_{m,f}\) according to the dependency structure learned from the source domain. For a GPR model, this posterior is a normal distribution, which is directly obtained from the closed-form solution shown in Eq. \ref{gpr} and collaboratively computed in the previous phase. 

To evaluate how well the observed value \(\tilde{x}_{m,f}\) fits the predicted distribution, we apply the confidence measure proposed by Jalali and Pfeifer \cite{jalali2016interpretable}:

\begin{equation}
   c_f(\tilde{x}_m) = 2 \Phi \left( \frac{|\tilde{x}_{m,f} - \mu_{g_f}(\tilde{x}_{m,\neg f})|}{\sigma_{g_f}(\tilde{x}_{m,\neg f})} \right)
   \label{confidences_big}
\end{equation}

where \(\Phi\) denotes the cumulative distribution function of a standard normal distribution. The terms \(\mu_{g_f}\) and \(\sigma_{g_f}\) correspond to the mean and standard deviation of the normal distribution predicted by the GPR model \(g_f\), respectively. This confidence score represents the probability that a value as extreme as \(\tilde{x}_{m,f}\), or more, relative to the predicted mean \(\mu_{g_f}(\tilde{x}_{m,\neg f})\), occurs within the posterior distribution predicted by \(g_f\).

The overall confidence for feature \(f\) in the target domain is then defined as the average of \(c_f(\tilde{x}_m)\) across all target domain samples:

\begin{equation}
    c_f = \frac{1}{n_t} \sum_{i=1}^{n_t} c_f(\tilde{x}_m)
    \label{confidences}
\end{equation}

Where \(n_t\) represents the total number of samples in the target domain. For each feature, \(c_f\) quantifies how well the source-domain dependencies for feature \(f\) align with those in the target domain. Once the confidence scores for all features have been computed, the target client then computes the weight of feature \(f\) as follows:

\begin{equation}
    w_f = (1 - c_f)^k
    \label{weight_formula}
\end{equation}

Here, \(k\) is a hyper-parameter specified by the target client, with \(k > 0\). This hyper-parameter determines how the confidence scores are transformed into feature weights. As \(k\) increases, progressively higher penalties are applied to features with low confidence, while features with higher confidence are penalized less severely.

In our experiments, we empirically evaluate the performance of our framework with respect to the weighting parameter \(k\) and adjust its value accordingly (Sections \ref{external_parameters} and \ref{experiments}).

\subsection{Federated Weighted Elastic Net Training}

The remaining phases of our framework involves collaboratively training weighted elastic nets in a federated manner, to preserve the privacy of individual source clients. By using a weighted elastic net, source clients scale the contribution of each feature in their local data to the regularization term based on the feature weights computed by the target client in the previous step.

The weighted elastic net solves the following optimization problem:

\begin{equation}
    \hat{\beta} = \underset{\beta}{\text{argmin}} \left( \| y - X\beta \|^2 + \lambda J(\beta) \right)
    \label{wen}
\end{equation}

where \( \| y - X\beta \|^2 \) represents the residual sum of squares on the source domain data, \( \lambda \) is the regularization parameter, and \( J(\beta) \) is the regularization term defined as:

\begin{equation}
    J(\beta) = \alpha \sum_{f=1}^F w_f |\beta_f| + \frac{1}{2}(1-\alpha) \sum_{f=1}^F w_f \beta_f^2
    \label{regularization_term}
\end{equation}

Training a weighted elastic net is essentially the same as training any other neural network in a federated setting \cite{mcmahan2017communication}. A weighted elastic net can be represented as a neural network with a single layer, where the coefficient for each feature is multiplied by its corresponding weight. Since the weights are fixed during the training process and all source domain owners have access to the same weights, the federated training primarily involves the iterative update of the model coefficients. This process is performed using secure aggregation \cite{bonawitz2016practical}, ensuring that local model updates remain protected from the aggregator at each iteration.

As shown in Equations \ref{wen}, \ref{regularization_term}, \textit{freda} has two external parameters: the proportion of \(L_1\) and \(L_2\) penalties in the weighted elastic net \(\alpha\), and the regularization parameter \(\lambda\). Following Handl et al. \cite{wenda}, we fix \(\alpha = 0.8\). The most challenging hyper-parameter to determine is the regularization parameter \(\lambda\), for which we adopt the prior knowledge approach proposed in \cite{wenda}.

\subsubsection{Optimal Lambda Prediction} \label{pn}

The regularization parameter \(\lambda\) determines the contribution of the regularization term \(J(\beta)\) to the overall objective function of the model. In our case, since domain adaptation heavily relies on this regularization term, selecting an optimal value for \(\lambda\) is critical. If \(\lambda\) is too small, the penalty on the features becomes weak, and differences among feature weights may not significantly influence the model's objective function. On the other hand, if \(\lambda\) is too large, the redistribution of coefficients across features with different weights may overly dominate the objective function, preventing the model from learning meaningful representations from the source domain \cite{wenda}.

Traditionally, the optimal value for the regularization parameter \(\lambda\) is determined through cross-validation. However, since our main objective is to perform unsupervised domain adaptation, using cross-validation to select the optimal \(\lambda\) is counter-intuitive in this context, as cross-validation cannot be performed on the target domain due to the absence of labels and must instead be performed on the training (source) data. In light of this, we adopt the prior knowledge approach proposed in \cite{wenda}.

The prior knowledge approach requires that side information about the similarities between the domains in the target domain is known by the target client. In \textit{freda}, this approach involves the target client partitioning the indices \(\{1, \cdots, l\}\) of all domains available in the target domain data \(X^T\), creating two subsets \(X^{t_1}\) and \(X^{t_2}\). Afterwards, source clients federatively train separate weighted elastic nets, where during the training of each model the \(\lambda\) value is selected by varying different possible values on a grid. Then, all these models trained with different values of \(\lambda\) are sent to the target client, where the target client picks the model that leads to the lowest MAE on the target domains in \(X^{t_1}\), assuming that the corresponding labels are available. Next, the target client fits a simple linear model for the relationship between domain similarity and the optimal \(\lambda\) values obtained from the weighted elastic nets provided by the source clients. Using this model, the target client predicts the optimal \(\lambda\) values for all domains in the remaining partition \(X^{t_2}\). These predicted values are then sent to the source domain owners for the final phase.

\subsubsection{Final Adaptive Model Training}

The source clients federatively train the final weighted elastic net models by selecting the regularization parameter \(\lambda\) based on the predicted values received from the target client in the previous step.

The model obtained at the end of this step is sent to the target client by the aggregator for the final prediction task on the target domain data.

\subsection{Implementation}

We implemented our framework in Python 3.8.18, the source code to reproduce the experiments is available on GitHub (\url{https://github.com/mdppml/FREDA}). For the feature models, we implemented our own GPR models and used the Python package GPy \cite{gpy2012} to compute the optimal values for the hyper-parameter optimization explained in Section \ref{hp_opt}. As for the weighted elastic nets, we used TensorFlow 2.13.1 with custom kernel regularization.

\section{Security Analysis}

We consider the security guarantees of our framework based on the following assumptions:
\begin{enumerate}
    \item The aggregator is semi-honest, meaning it follows the protocol but may attempt to infer sensitive information from observed data.
    \item Participating data owners are semi-honest, meaning they follow the protocol steps but may try to infer sensitive information from observed data or intermediate results.
    \item The communication channel between the aggregator and the clients is secure. It is not possible for other clients to perform man-in-the-middle attacks or eavesdrop on the exchanged messages.
\end{enumerate}

The assumption of a semi-honest aggregator and semi-honest clients is widely regarded as a standard in privacy-preserving machine learning \cite{aaai2023, icml2021, icml2022}. Based on these assumptions, we analyze the information which can be inferred by the participants during each phase of our framework.

\subsection{Security of the Federated Feature Model Training}

The training of feature models consists of two key steps (Section \ref{feature_models}): federated hyper-parameter optimization and federated GPR training.

\subsubsection{Security of the Federated Hyper-parameter Optimization}

This step is performed collaboratively between the source clients and the aggregator. The local data of the source clients are never exposed directly to either the aggregator or other data owners during this process. The only information exchanged during this step includes the optimal values of the prior variance on the coefficients $\sigma_p^2$ and the variance of the additive noise $\sigma_n^2$ (from the closed-form solution in Eq. \ref{gpr}). These values are computed locally by each source client for each feature.

The use of secure aggregation, implemented via zero-sum masking as formally proven in \cite{bonawitz2016practical}, ensures that the semi-honest aggregator cannot infer individual hyper-parameter values for any source client. Instead, the aggregator only gains access to the aggregated sum of hyper-parameters for each feature, which contains no exploitable information about the local datasets.

\subsubsection{Security of the Federated GPR Training} \label{gpr_security}

This step is performed collaboratively between all participants, which include the target client, source clients, and the aggregator.

\noindent \textbf{\textit{Aggregator's Perspective.}} \quad The information disclosed to the aggregator is limited to the number of local data samples held by each source client, as indicated by the size of each sub-matrix (Section \ref{gpr_computation}). This information is necessary for the correct partitioning of the intermediate mean matrix \(K_*K^{-1}\), but it does not reveal any details about the content of the local data samples. Due to the masking process employed, the aggregator cannot infer any information regarding the local datasets of the source or target clients. The only data accessible to the aggregator is the resulting Gram matrix \(G_{pq}\), which is computed from the masked matrices. The security of this masking process, implemented within the FLAKE framework, is formally proven in \cite{hannemann2023privacy}.

\noindent \textbf{\textit{Source Clients' Perspective.}} \quad The information available to each source client is limited to the respective sub-matrix they receive from the aggregator, which is derived from \(K_*K^{-1}\). The aggregator masks the intermediate matrix product by multiplying it with a mask matrix \(C\) (Section \ref{gpr_computation}), resulting in the masked matrix \(C K_*K^{-1}\). This masked matrix is then split into sub-matrices and sent to the respective source clients. Since each source client only has access to their own sub-matrix, they cannot infer any information about the overall matrix or the data of other clients without having access to the masking matrix \(C\).

\noindent \textbf{\textit{Target Client's Perspective.}} \quad The information available to the target client includes the masked mean vectors sent by the source clients, the inverse of the masking matrix \(C^{-1}\), and the predicted variance \(K_{**} - K_* K^{-1} K_*^\top\) sent by the aggregator. To remove the mask on the mean vectors, the target client computes:
\[
K_* K^{-1} y = C^{-1} \sum_{i=1}^N (C K_* K^{-1})^{s_i} y^{s_i}.
\]
Since the individual vectors from the source clients are aggregated during this process, the target client cannot infer any meaningful information about the local datasets of the source clients. Regarding the predicted variance \(K_{**} - K_* K^{-1} K_*^\top\), although the target client has access to \(K_{**}\), it can compute \(- K_* K^{-1} K_*^\top\). However, without access to the individual data matrices of the source clients, the target client cannot infer any meaningful information, as there are infinitely many possible source data matrices that would result in the same \(K\) and \(K_*\) \cite{hannemann2023privacy}.

\subsection{Security of the Feature Weight Computation}

This step is performed locally by the target client. At the end of this process, the target client sends the computed feature weights for all tissues to the aggregator. The aggregator then distributes these weights to the source clients.

The local data of the target client is never directly exposed to either the aggregator or the source clients during this process. The only information shared with the aggregator and source clients consists of the feature weights for each tissue in the target domain. These feature weight vectors are aggregate values, computed by the target client by averaging the feature weights of the samples belonging to each tissue within the target domain data (Section \ref{confidence_score_computation}). Consequently, it is not possible for the aggregator or the source clients to infer any meaningful information on the target data.

\subsection{Security of the Optimal Lambda Prediction} \label{lambda_security}

This step is performed collaboratively among all participants, including the target client, source clients, and the aggregator. Throughout this process, the local data of the target client and the source clients are never directly exposed to either the aggregator or any other participant.

\noindent \textbf{\textit{Aggregator's Perspective.}} \quad During the federated training process, the aggregator receives masked model updates from the source clients, which it aggregates to compute the coefficients of the global model for the next iteration. The use of secure aggregation, implemented via zero-sum masking as formally proven in \cite{bonawitz2016practical}, ensures that a semi-honest aggregator cannot infer individual updates from any source client.

After the target client predicts the optimal lambda values, the aggregator receives these values. Since the target data is never directly shared, the aggregator cannot infer any information about the target data from the optimal lambda values.

\noindent \textbf{\textit{Source Clients' Perspective.}} \quad During the federated training process, source clients receive the updated global model after each training round. This global model consists only of model coefficients, ensuring that no information about other source clients' local data is exposed. Source clients never directly access one another's data, preserving the privacy of all local source data.

Once the target client predicts the optimal lambda values, the aggregator distributes these values to the source clients. Since the target data is never directly shared, source clients cannot infer any information about the target data from the received lambda values.

\noindent \textbf{\textit{Target Client's Perspective.}} \quad The target client receives only the coefficients of models trained using various regularization parameters. These model coefficients are aggregate values that do not reveal any specific information about the local data of any source client.

\subsection{Security of the Final Adaptive Model Training}

This step is performed collaboratively among all participants, including the target client, source clients, and the aggregator. Throughout this process, the local data of the target client and the source clients are never directly exposed to either the aggregator or any other participant.

\noindent \textbf{\textit{Aggregator's Perspective.}} \quad The aggregator receives masked model parameters from the source clients, which it aggregates to compute the coefficients of the global model for the next iteration. The use of secure aggregation, implemented via zero-sum masking as formally proven in \cite{bonawitz2016practical}, ensures that a semi-honest aggregator cannot infer individual updates from any source client.

\noindent \textbf{\textit{Source Clients' Perspective.}} \quad During the federated training process, source clients receive the updated global model after each training round. This global model consists only of model coefficients, ensuring that no information about other source clients' local data is exposed. Source clients never directly access one another's data, preserving the privacy of all local source data.

\noindent \textbf{\textit{Target Client's Perspective.}} \quad The target client receives only the coefficients of models trained using various regularization parameters. These model coefficients are aggregate values that do not reveal any specific information about the local data of any source client.

\section{Conclusion}

In this article, we introduce \textit{freda}, the first method to enable privacy-preserving, federated unsupervised domain adaptation for regression tasks in high-dimensional, small-scale biological datasets. Unlike existing approaches that rely on deep learning or centralized training, \textit{freda} allows multiple entities to collaboratively model complex feature relationships while maintaining complete data privacy. By leveraging a novel combination of randomized encoding and secure aggregation, our framework overcomes the fundamental challenge of training Gaussian Processes in federated settings by solving the need for centralized pairwise computations on data that cannot be directly shared. 

Gaussian Processes are well-suited for modeling high-dimensional, small-sample datasets due to their probabilistic nature and ability to provide uncertainty estimates. However, their reliance on full data access has historically made them impractical for distributed learning scenarios. \textit{Freda} is the first method to overcome this limitation, demonstrating that federated training of Gaussian Processes is both feasible and effective in real-world scenarios where privacy constraints and data heterogeneity present significant challenges.

To evaluate \textit{freda}, we applied it to a benchmark task of age prediction from DNA methylation data, comparing its performance against the state-of-the-art centralized method \textit{wenda} and the non-adaptive baseline \textit{en-ls}. Our results show that \textit{freda} achieves comparable performance to the state-of-the-art while ensuring complete privacy and enabling cross-institutional collaboration without direct data sharing. Our findings highlight the potential of \textit{freda} as a generalizable framework for domain adaptation in high-dimensional biological datasets.

By introducing the first privacy-preserving, federated training framework which utilize Gaussian Processes in domain adaptation, this work provides a foundation for secure, data-efficient modeling in computational biology and other fields where data privacy, heterogeneity and sample volume are major concerns.

\newpage

\bibliography{sn-bibliography}

\end{document}